\documentclass[10pt, conference, compsocconf]{IEEEtran}
\ifCLASSINFOpdf
   \usepackage[pdftex]{graphicx}
   \DeclareGraphicsExtensions{.pdf,.jpeg,.png}
\else
   \usepackage[dvips]{graphicx}
\fi
\usepackage[cmex10]{amsmath}
\begin{document}
%
\title{Using Soft Computer Techniques on Smart Devices for Monitoring Chronic Diseases: the CHRONIOUS case}


\author{%
Piero Giacomelli \\
Tesan S.p.A. \\
email:giacomelli@tesan.it \\
 \and
Giulia Munaro\footnote{munaro@tesan.it} \\
Tesan S.p.A. \\
email:munaro@tesan.it \\
 \and
Roberto Rosso\footnote{rosso@tesan.it}   \\
Tesan S.p.A. \\
email:rosso@tesan.it \\
           }

%


\maketitle

\begin{abstract}
		CHRONIOUS is an Open, Ubiquitous and Adaptive Chronic Disease Management Platform for Chronic Obstructive Pulmonary Disease(COPD)
	Chronic Kidney Disease (CKD) and Renal Insufficiency. It consists of several modules: an ontology based literature search engine, a rule based decision support system, remote sensors interacting with lifestyle interfaces (PDA, monitor touchscreen) and a machine learning module. All these modules interact each other to allow the monitoring of two types of chronic diseases and to help clinician in taking decision for cure purpose.
	This  paper illustrates how some machine learning algorithms and a rule based decision support system can be used in smart devices, to  monitor chronic patient. We will analyse how a set of machine learning algorithms can be used in smart devices to  alert the clinician in case of a patient health condition worsening trend.
\end{abstract}


%
\IEEEpeerreviewmaketitle

\section{Introduction}

Scientific advances over the past 150 years, particularly in the medical field, have allowed the extension of life expectancy in western countries and this trend seems to increase in future years. Conservative estimates suggest that by 2030 in EU countries the proportion of people over 60 years  regard the entire population will be around 50$\%$; this means that we will see a gradual increase in the number of those subjects with chronic diseases (ie diseases not involving healing), that will therefore increase the cost and effort over health care facilities \cite{CarmineZoccali06012010}.
As consequence of the exponential growth of hardware and software infrastructure it is possible to rethink the whole approach to the treatment of complex chronic disease, by limiting the hospitalization only to a severe worsening of patient's condition. This was the original idea behind the CHRONIOUS project: constructing a generic platform to monitor, in an unobtrusive way, a chronic disease patient with two goals\cite{Vitacca01022009}:
\begin{itemize}
\renewcommand{\labelitemi}{$\bullet$}
\item{Improve the patients quality of life, by reducing as much as possible the hospitalizations.}
\item{Allow the clinician a continuous monitoring the patients, both in standard and  potential risk situations.}
\end{itemize}
\indent
To gain this two goals, the CHRONIOUS platform has to integrate different technologies and hardware and software modules that need to interact among themselves. This  paper is organized as follows: in the first section, the general structure of CHRONIOUS hardware and software modules are described. A deep analysis of the preprocessing algorithms covers the entire second section. Section three is dedicated to illustrate the machine learning algorithms. In last section we will evaluate possible improvements to the Chronious intelligence system.

\section{The CHRONIOUS system: an overview}
CHRONIOUS system deals with COPD and CKD: for these different types of chronic diseases, \hyphenation{ac-cor-ding}according to the medical guidelines, it is important to monitor different data in a way to check patient health status and to activate suitable emergency alarms for the clinician (for the COPD: ECG, SPO$2$ and respiratory rate; for CKD:  glucose level, body weight and blood pressure) in case of critical event.

The CHRONIUOS platform consists of many modules that act together.

\begin{figure}[htbp]
	\centering
		\includegraphics[width=0.5\textwidth]{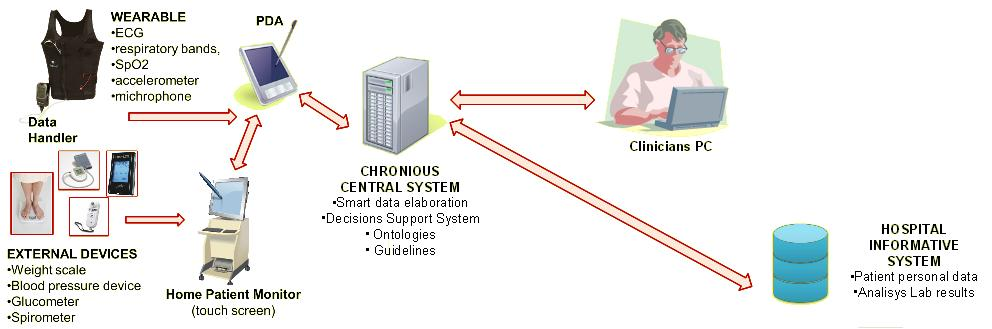}
	\caption{Chronious modules}
	\label{}
\end{figure}
We can organize them in three frameworks
\begin{itemize}
\renewcommand{\labelitemi}{$\bullet$}
\item{Patient Sensor Framework}
\item{Communication Framework.}
\item{Monitor Framework.}
\end{itemize}
\indent
The Patient Sensor Framework consists of hardware devices used to grab data from the patients. The hardware equipments, installed at patient's home, are
\begin{itemize}
\renewcommand{\labelitemi}{$\bullet$}
\item{Wearable and external devices.}
\item{Touch screen Home Patient Monitor(HPM).}
\item{Personal Digital Assistant (PDA).}
\end{itemize}
\indent
The data collected are of two types:
\begin{itemize}
\renewcommand{\labelitemi}{$\bullet$}
\item{silent, when the data recording is automated and it does not involve the patient interaction, as respiratory frequency measurement for COPD patients from the wearable device}
\item{no-silent, when it is requested a direct patient or caregiver interaction with the device, as for the blood pressure and questionnaires inputted by the HPM for diet, activity and food intake monitoring. The no-silent data acquisition is particularly important  for monitoring CDK patient lifestyle.
}
\end{itemize}
\indent
 During the day there are several measureaments, with different time intervals and different frequencies; only one data transmission is done, if there is no worsening in patient parameters. For COPD patient all the data are collected in a silent mode from a wereable t-shirt and trasmitted to the PDA via Bluetooth; for CDK patient all the measurements, including a lifestyle questionnarie are stored in the HPM and trasmitted in silent mode to the PDA.
PDA is in charge of doing the following action
\begin{itemize}
\renewcommand{\labelitemi}{$\bullet$}
\item{Collect all the data from the devices.}
\item{Use a set of machine learning algorithms to determinate if it is needed to force a non scheduled trasmission to the Monitoring Framework.}
\item{Trasmit the collected and analysed information to the Monitoring Framework and receive back the changes, that will affect the interaction between the patient and the other devices.}
\end{itemize}
\indent

The Communication Framework is in charge of transmitting data among \hyphenation{dif-fe-rent} devices and from PDA to Central DB. This transmission is done using messages in a predefined xml format. The device that is in charge of doing the transmission of the xmls is the PDA.
\newline
\indent
The Monitor framework is the principal interface between the clinician and the patient. It receives data from the remote PDAs and it transmits back the therapy for food intake, drug intake, activities and scheduled measurements. The PDA is equipped also with a rule based decision support system, that contains an updatable set of rules created from the literature and validated by clinicians.

The rule based decision support system is in charge of analysing the data arrived from the PDAs and deciding if there is a grave or mild worsening of patient's conditions.
It is also able to alert the clinician and propose the suggestions about the action to do in several cases.
In the next section we will analyse the preprocessing phase needed to activate the intelligence in the PDA.

\section{The PDA CHRONIOUS preprocessing phase}
As we pointed out above, the Personal Digital Assistant is a smartphone equipped with WINDOWS MOBILE 6.5 Operating system, a SQL SERVER 2005 COMPACT database and a .NET FRAMEWORK 3.5.
The data registered by the PDA are the following
\begin{itemize}
\renewcommand{\labelitemi}{$\bullet$}
\item{Data from the wereable jacket.}
\item{Answers to questionnaries concerning dietary habits, drug intake and  lifestyle of the patient.}
\item{Data from the home patient monitor like blood pressure, glucometer measurements and body weight.}
\end{itemize}

\indent
Once these data are collected they are saved to the PDA database and a set of algorithms are triggered to analyse these data.
Since the PDA analyzes data for  two different diseases, two sets of different algorithms are used.
The fundamental data needed by COPD treatment are ECG signals and Respiration data, so in case of a COPD patient we have a first processing Electrocardiogram Pre-processing. After that a Feature extraction phase is needed and at the end an Evaluation phase of the extracted features is done to determinate if an alert must be triggered to the Central System.
For external devices used in particular by CDK patients, there is no need of a preprocessing phase beacuse foundamental measures are the ones provided by the glucometer, the weight scale and the blood pressure measure; so they are discrete time and directly used by the set of machine learning. Combined with these data, the answers to a set of queries  concerning food intake, drug intake, lifestyle and mental status are passed to a set of machine learning algorithms to evaluate the whole patient condition.
In next subsections we will analyse first the COPD set of algorithms used.
\subsection{Preprocessing of COPD signals}
The aim of Preprocessing Phase is to improve the general quality of the ECG, for more accurate analysis and measurement, because there's the possibilty to have some noises on the signals.
Possible noises in the signal include

\begin{itemize}
\renewcommand{\labelitemi}{$\bullet$}
\item{Low frequency Base Line Wandering (BW) caused by respiration and body movements.}
\item{High frequency random noises caused by mains interference (50 or 60Hz).}
\item{Muscular activity and random shifts of the ECG signal amplitude caused by poor electrode contact and body movements.}
\end{itemize}
\indent
The preprocessing comprises:
\begin{itemize}
\renewcommand{\labelitemi}{$\bullet$}
\item{Removal of base line wandering.}
\item{Removal of high frequency noise.}
\item{QRS detection.}
\end{itemize}
\indent

The BW which is is an extragenoeous low-frequency activity which may interfere with the signal analysis, rendering its clinical interpretation inaccurate and misleading.
Two major techniques are employed for BW removal:
\begin{itemize}
\renewcommand{\labelitemi}{$\bullet$}
\item{Linear filtering: involves the design of a LTI high pass filter with
cut off in way that the clinical information in the ECG is preserved
and  the BW is removed as much as possible.}
\item{Polynomial fitting: includes the fitting of polynomials to
representative points (knots) in the ECG, with one knot for each
beat. Knots are selected from a silent segment, e.g. the PQ
interval. A polynomial is fitted so that it passes through every
knot in a smooth fashion.}

\end{itemize}
\indent

The High Frequency Noise can be caused by the high frequency as well
as power supply interference from the ECG signal. It's removal  is done using:

\begin{itemize}
\renewcommand{\labelitemi}{$\bullet$}
\item{The Daubechies (DB4) wavelet employed on the
basis of the resemblance and similar frequency
response characteristics of the db4 basis function
with the ECG waveform}.
\item{ Using wavelets to remove noise from a signal
requires identifying which components contain the noise,
using optimal methods to threshold them, and then reconstructing
the signal using the thresholded coefficients.}
\end{itemize}
\indent

The prepocessing phase finally deals with the QRS detection. The main features that should be calculated: the Inter-beat (RR) interval and the Heart Rate Variability (variation in the beat-to-beat interval).
For the Inter-beat (RR) interval, two methods have been explored
\begin{itemize}
\renewcommand{\labelitemi}{$\bullet$}
\item{Filtering the ECG signal with continuous (CWT) and fast wavelet\cite{Chesnokov}
transforms (FWT)\footnote{The reconstructed ECG signal after denoising contains only spikes with non-zero
values at the location of QRS complexes. From this signal, the PQ junction and J point
can be located as the boundaries of the spike. If the length of the spike is more or
less than a predefined QRS length range it is annotated as noise and if the voltage is
below a certain threshold, it is annotated as an artifact.
The next stage is the detection of the T wave, and the P wave in the PQ interval. The
peaks of Q, R and S waves are identified in the annotated part of the ECG signal from
the PQ junction to J point.}.}
\item{Following Pan-Tompkins\cite{Pan}, wavelets are used to remove noise from a signal
requires identifying which component or
components contain the noise, using optimal
methods to threshold them, and then reconstructing
the signal using the thresholded coefficients\footnote{
The algorithm includes a series of filters and
methods that perform lowpass, high-pass, derivative, squaring, integration, adaptive
thresholding and search procedures.
}.}
\end{itemize}
\indent

All the previuos features are extracted from ECG signals. For COPD patient also the Respiratory Rate is a fundamental parameter that need to be analysed.
In order to calculate the respiration rate using the reference respiration signal, a dominant frequency detection algorithm, based on short-time Fourier transform (STFT) \cite{0967-3334-26-5-R01}, is applied.

The STFT is a localized Fourier transform, utilizing a Hamming window:
\begin{equation}
STFT(f(t)) = STFT(\omega, \tau) = \int_{\infty}^{\infty} f(t) w(t-\tau) e^{-j\omega t} dt
\end{equation}

were $w(t)$ is the window function, commonly a Hann window or Gaussian hill centered around zero, and f(t) is the signal to be transformed.
Because frequency components of the respiration signal are very low (<$2$Hz), a window size of $60$ seconds is selected.Every 60s, the hamming window is multiplied to the respiration
signal, and the result is transformed to the frequency domain using
Fourier transform. The dominant frequency is then detected by
finding the maximum amplitude of the spectrums. When the
dominant frequency components are found, inverse numbers are
calculated in order to obtain the respiration rate.
After this first preprocessing phase for COPD patients we wil now analyse the which kind of Features are extracted.

\subsection{Features Extraction for COPD patients }
From the Inter-beat (RR) interval and the Heart Rate Variability,
several features can be extracted, either in time or in frequency
domain.
\newline
\indent
Dealing with Time domain the values extracted are
\begin{enumerate}
\item{ SDNN(msec): Standard deviation of all normal RR intervals in
the entire ECG recording using the following
\begin{equation}
sdnn = \sqrt{\frac{1}{n} \sum_{i=1}^n(NN_i - m)^2 }
\end{equation}
where $NN_i$ is the duration of the $i$-th $NN$ interval in the analyzed
ECG, $n$ is the number of all $NN$ intervals, and $m$ is their mean duration.
}
\item{
SDANM(msec): Standard deviation of the mean of the normal
RR intervals for each $5$ minutes period of the ECG recording.
}
\item{
SDNNIDX (msec): Mean of the standard deviations of all
normal RR intervals for all $5$ minutes segments of the ECG
recording.
}
\item{
pNN50 (
intervals that are greater than $50$ msec, computed over the
entire ECG recording.
}
\item{
r-MSSD (msec): Square root of the mean of the sum of the
squares of differences between adjacent normal RR intervals
over the entire ECG recording the formula is
\begin{equation}
rMSSD = \sqrt{\frac{1}{n-1} \sum_{i=1}^n (NN_{i+1}-NN_{i})^2}
\end{equation}
where NNi is the duration of the $i$-th $NN$ interval in the analyzed
ECG and $n$ is the number of all $NN$ intervals.
}
\end{enumerate}

If we now move to the frequency domain the Feature Extraction on the PDA studies two bands:
\begin{enumerate}
\item{ The Low – Frequency band (LF), which includes frequencies in
the area $[0.03 – 0.15]$ Hz.
}
\item{
The High – Frequency band (HF), which includes frequencies in
the area $[0.15 – 0.40]$ Hz.
}
\end{enumerate}
If we now move to the Respiration signal several features can be extracted either directly or indirectly we focused on:
\begin{enumerate}

\item{ Respiration Rate: The number of breaths per minute.}
\item{Tidal Volume (VT): The normal volume of the air inhaled after an
exhalation.}
\item{Vital capacity (VC): The volume of a full expiration. This metric
depends on the size of the lungs, elasticity, integrity of the
airways and other parameters, therefore it is highly variable
between subjects.}
\item{Residual volume (VR): The volume that remains in the lungs
following maximum exhalation.}
\end{enumerate}

After all the preprocessing phase of the data gathered by the weareable devices, all these information are passed to a Classification System. The classification system is responsible for the analys of  the outcome from the preprocessing phase for the COPD patients and of the data gathered by the external devices and questionnaries inputted by the patient himself.
Below we will see how  the software is used to transform all these rich set of data in an information to be, in case, transmitted real-time or scheduled to the Monitoring Framework.

\section{The CHRONIOUS Classification System}
After the collected data have been preprocessed for COPD patient and all the CKD patient input have been acquired, a set of machine learning algorithms are fired up to decide if a potentially risky situation is present. The aim of these tools is alerting the Central System that contains a rule based decision support system, for a better evaluation of the message triggered by the PDA. In case the message containg an alarm for life risk danger, the Central Decision Support System is able to alert the emergency staff or to suggest the clinician to modify the therapy approach.
Most of these tools need a preprocessing phase for identifying the correct parameters that need to be validated by the clinicians. This means that in the first validation of the CHRONIOUS project a large amount of efforts has been dedicated to gather feedback from the clinicians about the correctness of the rules / parameters that have been inferred by the algorithms. In these phase another important effort has been dedicated by the technicians to evaluate some probability index for  fuzzy data measures.

The CHRONIOUS Classification System is composed of the following part:

\begin{itemize}
\renewcommand{\labelitemi}{$\bullet$}
\item{A light rule based expert system.}
\item{A supervised classification system.}
\end{itemize}
\indent

The light rule based expert system is an xml parser that is able to extract from an xml a set of "`if then"' rules created an validated by the clinician.
With these rules  combined with the data collected, the rule base system is able to decide if a patient is in a potentially life-risk situation.
For example the following rules will generate an immediate alarm to the central system:
\begin{itemize}
\renewcommand{\labelitemi}{$\bullet$}
\item{
The Hearth Rate is above 120 bpm for both COPD and CKD patient.
}
\item{
If the weight increase by $2\%$ in the last 24 hour for CDK patient .
}
\end{itemize}
\indent
These rules are most for alarm triggering. It means that they aren't use for light monitor alerting. In the CHRONIOUS PDA system the Supervised Classification System is composed of the following machine algorithms:

\begin{itemize}
\renewcommand{\labelitemi}{$\bullet$}
\item{Support Vector Machines \cite{Cristianini}}
\item{Random Forest \cite{Brieman}}
\item{Multi-Layer Perceptron \cite{Haykin}}
\item{Decision Tree \cite{Quinlan}}
\item{Na\"ive Bayes \cite{Friedman}}
\item{Partial decision Trees \cite{Eibe}}
\item{Bayesian Network \cite{holmes}}
\end{itemize}
\indent

Apart from Bayesian Network, the other algorithms have been trained with a dataset to generate a set of rule that have been validated by clinician. The Bayesian Network have been used to identified possible rules about the mental and stress evaluator of the patient.
This means that once trained, the rules generate can be used to identify is some stressfull condition can alter some parameter and leading to a worsening of the general state of the patient.
The use of these algorithms was needed beacuse for the CKD patient the diet covers the most part of the medical treatment, so any factor that can influence a changement on the diet intake, would potentially and indirectly lead to a worsening of the patient condition. For example, if a female CKD patient feeld sad, these condition could lead her to eat a bigger piece of pie for satisfatcion purporse. In general a bad feeling could lead chronic patients to be uncompliance with the medical treatment. However the kind of rules aren't liked the vital signs so their fuzzyness could be identified by these type of algorithms. Clearly while diet is important  to avoid worsening on CDK patient conditions, on the other side the lifestyle could be and indirect cause of COKD worsening condition.
Nevertheless for both of them we need static rules to have alarms sending, because for both for example having a body temperature above $38$ could be a risky situation were an hospitalization is needed for both COPD and CKD patients.
Considering the training dataset a set of $41$ attributes have been identified.
These data comes from the $2$ sets of $2$ hours health recordings ($11$ attributes), food input module ($12$ attributes), drug intake module ($1$
attribute), activity input module ($2$ attributes), questionnaire ($13$ attributes) and external device ($2$ attributes).
Some of the results of some classifier are shown in table $1$.

\begin{table}[htbp]
	\centering
	\caption{Classification System: 	 MAE: Mean absolute error, RMSE: Root mean squared error, RAE : Relative absolute error, CI: Correctly/Instances}
\begin{tabular}{|c|ccc|r|}
	\hline
	 Method &  MAE   &  RMSE & RAE  & CI  \\
	\hline
PART & 0.1336 & 0.312 &  57.81 \% &  2.67  \\
J48  & 0.1336 & 0.321 & 57.81 \% & 2.67 \\
Forest & 0.2 & 0.341 & 86.52 \% & 1.75 \\
Na\"ive & 0.127 & 0.343 & 54.95 \% & 2.67 \\
	\hline
\end{tabular}

	\label{tab:table1}
\end{table}
Again we point out that even if there are some errors due to false positive matches, the PDA system in these case would only generate a rule that will send a message to the clinician that most of the times would only say that a light worsening is present. The core intelligence that deals with the central system would be the real suggestion system that would indicate to clinician a suggestion on how to act to possibly revert the trend.
The Bayesian Network is the fundament algorithm for the Mental support tool. It uses a set of attributes that affect a stress index and they weight based on the clinician's feedback as shown in table of Figure 2. When the total stress indicator is above a certain value a light alarm is triggered to the Central Database to inform clinician of a potentially worsening of patient conditions.
In the same way a module in the PDA is in charge of the rules concerning the lifestyle od the patient: the lifestyle tool. It collects data inputted by the patient or caregiver in a validation phase and using a Bayesian Network is able to compute an index of good or poor lifestyle of the patient also in this case if the poor lifestyle is find a light alarm is sended to the clinician for monitoring purporse.

\begin{figure}[ht]
	\centering
		\includegraphics[width=0.50\textwidth]{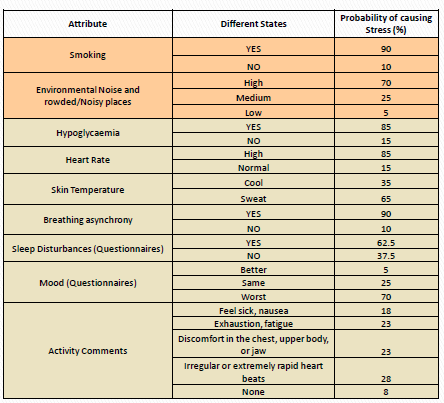}
	\caption{Attributes}
	\label{fig:mental}
\end{figure}

\section{Conclusions an improvements}
In these paper we presented a set of machine learning algorithms store in a smart phone that, combined with some external devices and patient inputted data, can be used for a first monitor/alert system for treatment of patient affected by chronic diseases.
Dealing with telemedicine application these kind of software could help to improve patient  life quality and could be also be a valid help for clinician to allow a more precise monitoring of patient conditions without need of the physical presence of the clinician. Apart these potentially advantage a PDA equipped with these kind of application can suffer of some limitations.
During developing phase we face these problems:
\begin{itemize}
	\item{Heavy resource consumption of preprocessing algorithms.}
	\item{Updating a trained supervised algorithm. }	
\end{itemize}
\indent
The preprocessing algorithm for COPD parameter denoising is the most memory/CPU consumption. This potentially can became a problem when we deal with  life risk situations,  because in the time that the algorithm denoise the ECG signal, the patient can became unconscius and so precious time can be lost in these phase. Other time is losed due to the huge amount of signal trasmitted from the wereable, this because we deal with an ECG signal that is composed of a mean of ten measurements per second so this means that 5 minute of ECG signal became nearly $3000$ sql commands to a SQL SERVER 2005 COMPACT that is not so performant in these case. Increasing hardware requirements of the PDA can be a first solution however it would be interesting to understand if a relational database is the best solution for storing these type of data or other structure would be more confortable for storing purpose.
The other important issue is that once the supervised learning algorithms are trained also a little change in one parameter will need to retrain the algorithms and most important, it would need a new validation of the outputs by the clinicians. These can lead to difficults when after these phase the new trained algorithms need to be updated on the pda, this because the Chronious Communication Framework is only able to transmits value from the PDA to Monitor Framework and back. In these case an  interesting solution could be also to allow remote updating of the structure validated. For example a trained neural network could read the weights matrix from a structure upgradable by the communication framework. Apart from these improvements, and many others that could lead to a software system closer as much as possbile to the clinician and patient needs,  it is our opinion that with the smart devices that are closer to a normal PC, the algorithms presented in this paper could become an important part of the telemedicine platforms.

\bibliographystyle{IEEEtran}
\bibliography{etelemed2011}

\begin{thebibliography}{10}
\providecommand{\url}[1]{#1}
\csname url@samestyle\endcsname
\providecommand{\newblock}{\relax}
\providecommand{\bibinfo}[2]{#2}
\providecommand{\BIBentrySTDinterwordspacing}{\spaceskip=0pt\relax}
\providecommand{\BIBentryALTinterwordstretchfactor}{4}
\providecommand{\BIBentryALTinterwordspacing}{\spaceskip=\fontdimen2\font plus
\BIBentryALTinterwordstretchfactor\fontdimen3\font minus
  \fontdimen4\font\relax}
\providecommand{\BIBforeignlanguage}[2]{{%
\expandafter\ifx\csname l@#1\endcsname\relax
\typeout{** WARNING: IEEEtran.bst: No hyphenation pattern has been}%
\typeout{** loaded for the language `#1'. Using the pattern for}%
\typeout{** the default language instead.}%
\else
\language=\csname l@#1\endcsname
\fi
#2}}
\providecommand{\BIBdecl}{\relax}
\BIBdecl

\bibitem{CarmineZoccali06012010}
C.~Zoccali, A.~Kramer, and K.~Jager, ``Chronic kidney disease and end-stage
  renal disease—a review produced to contribute to the report 'the status of
  health in the european union: towards a healthier europe','' \emph{NDT Plus},
  vol.~3, no.~2, pp. 213--224, 2010.

\bibitem{Vitacca01022009}
M.~Vitacca, L.~Bianchi, A.~Guerra, C.~Fracchia, A.~Spanevello, B.~Balbi, and
  S.~Scalvini, ``Tele-assistance in chronic respiratory failure patients: a
  randomised clinical trial,'' \emph{European Respiratory Journal}, vol.~33,
  no.~2, pp. 411--418, 2009.

\bibitem{Chesnokov}
Y.~Chesnokov, D.~Nerukh, and R.~Glen, ``Individually adaptable automatic qt
  detector,'' \emph{Computers in Cardiology}, no.~2, pp. 337--340, 2006.

\bibitem{Pan}
J.~Pan and W.~J. Tompkins, ``{A real-time QRS detection algorithm.}''
  \emph{IEEE Trans Biomed Eng}, vol.~32, pp. 230--236, 1985.

\bibitem{0967-3334-26-5-R01}
P.~S. Addison, ``Wavelet transforms and the ecg: a review,''
  \emph{Physiological Measurement}, vol.~26, no.~5, p. R155, 2005.

\bibitem{Cristianini}
N.~Cristianini and J.~Shawe-Taylor, \emph{An introduction to support vector
  machines and other kernel based learning methods}.\hskip 1em plus 0.5em minus
  0.4em\relax Cambridge University Press, 2000.

\bibitem{Brieman}
L.~Breiman, ``Random forests,'' \emph{Machine Learning}, vol.~45, pp. 5--32,
  2001.

\bibitem{Haykin}
S.~Haykin, \emph{Neural Networks and Learning Machines (3rd Edition)}.\hskip
  1em plus 0.5em minus 0.4em\relax Prentice Hall, 2008.

\bibitem{Quinlan}
J.~R. Quinlan, ``Learning decision tree classifiers,'' \emph{ACM Comput.
  Surv.}, vol.~28, pp. 71--72, 1996.

\bibitem{Friedman}
N.~Friedman, D.~Geiger, and M.~Goldszmidt, ``Bayesian network classifiers,''
  \emph{Machine Learning}, vol.~29, pp. 131--163, 1997.

\bibitem{Eibe}
E.~Frank and I.~H. Witten, ``Generating accurate rule sets without global
  optimization,'' in \emph{Proceedings of the Fifteenth International
  Conference on Machine Learning}, 1998, pp. 144--151.

\bibitem{holmes}
D.~Holmes and L.~C. Jain, \emph{Innovations in Bayesian Networks Theory and
  Applications}.\hskip 1em plus 0.5em minus 0.4em\relax Springer Netherlands,
  2008.

\end{thebibliography}

\end{document}